%% file: main.tex
\documentclass{article}

\PassOptionsToPackage{numbers, compress}{natbib}

\usepackage[final]{neurips_data_2021}

\usepackage[utf8]{inputenc} 
\usepackage[T1]{fontenc}    
\usepackage{hyperref}       
\usepackage{url}            
\usepackage{booktabs}       
\usepackage{amsfonts}       
\usepackage{nicefrac}       
\usepackage{microtype}      
\usepackage{xcolor}         
\usepackage{multirow}
\usepackage{graphicx}
\usepackage{listings}
\usepackage{amsmath}
\usepackage{pifont}
\usepackage{tabu}
\usepackage{longtable}
\usepackage{comment}
\usepackage{enumitem}
\usepackage{xspace}
\usepackage{pythonhighlight}

\definecolor{codegreen}{rgb}{0,0.6,0}
\definecolor{codegray}{rgb}{0.5,0.5,0.5}
\definecolor{codepurple}{rgb}{0.58,0,0.82}
\definecolor{backcolour}{rgb}{0.95,0.95,0.92}

\lstdefinestyle{mystyle}{
    backgroundcolor=\color{white},   
    commentstyle=\color{codegreen},
    keywordstyle=\color{magenta},
    numberstyle=\tiny\color{codegray},
    stringstyle=\color{codepurple},
    basicstyle=\ttfamily\footnotesize,
    breakatwhitespace=false,         
    breaklines=true,                 
    captionpos=b,                    
    keepspaces=true,                 
    numbers=right,                    
    numbersep=5pt,                  
    showspaces=false,                
    showstringspaces=false,
    showtabs=false,                  
    tabsize=2
}

\definecolor{jade}{rgb}{0.0, 0.66, 0.42}
\definecolor{cornellred}{rgb}{0.7, 0.11, 0.11}
\definecolor{bondiblue}{rgb}{0.0, 0.58, 0.71}
\renewcommand{\arraystretch}{1.5}

\hypersetup{
    colorlinks=true,
    citecolor=bondiblue,
    linkcolor=bondiblue,
    filecolor=magenta,    
    urlcolor=codegray,
}

\newcommand{\green}[1]{\textbf{\textcolor{jade}{#1}}}
\newcommand{\cmark}{\text{\ding{51}}}
\newcommand{\xmark}{\text{\ding{55}}}
\newcommand{\G}{\green{$\mathbb{G}$} }

\newcommand{\etal}{et al.\@\xspace}

\title{Full-Cycle Energy Consumption Benchmark\\ for Low-Carbon Computer Vision}

\author{%
  Bo Li$^{1}$ \; Xinyang Jiang$^{1}$ \; Donglin Bai$^{1}$ \;
  Yuge Zhang$^{1}$ \; Ningxin Zheng$^{1}$ \; \\
  \textbf{Xuanyi Dong}$^{2}$ \;  \textbf{Lu Liu}$^{2}$ \;
  \textbf{Yuqing Yang}$^{1}$ \; \textbf{Dongsheng Li}$^{1}$ \\
  $^1$Microsoft Research Asia \; $^2$University of Technology Sydney \\ 
}

\begin{document}

\maketitle

\begin{abstract}
The energy consumption of deep learning models is increasing at a breathtaking rate, which raises concerns due to potential negative effects on carbon neutrality in the context of global warming and climate change. 
With the progress of efficient deep learning techniques, e.g., model compression, researchers can obtain efficient models with fewer parameters and smaller latency. 
However, most of the existing efficient deep learning methods do not explicitly consider energy consumption as a key performance indicator. 
Furthermore, existing methods mostly focus on the inference costs of the resulting efficient models, but neglect the notable energy consumption throughout the entire life cycle of the algorithm. 
In this paper, we present the first large-scale energy consumption benchmark for efficient computer vision models, where a new metric is proposed to explicitly evaluate the full-cycle energy consumption under different model usage intensity. 
The benchmark can provide insights for low carbon emission when selecting efficient deep learning algorithms in different model usage scenarios.
\end{abstract}

\section{Introduction}\label{sec:intro}

Global warming and climate change have become the most pressing issues to the modern society, with already observable effects world-wide~\cite{field2012managing}, including shrunken glaciers, shifted plant and animal habitats, wild-fire, extreme weather, etc. 
For climate issues, artificial intelligence (AI) is a double-edged sword. On the one hand, it provides new technologies to control climate problems in many areas (such as smart-grid design, climate change prediction) \cite{rolnick2019tackling}. On the other hand, AI itself is also a significant carbon emitter. 
There has been a clear trend in AI community to develop larger deep learning models for better performance, and the increase rate of energy consumption of the state-of-the-art AI models is alarming.  For example, the number of parameters in state-of-the-art language models increased by $\sim$1870 times in about 3 years (from 93.6 Million \cite{peters2018deep} to 175 Billion \cite{NEURIPS2020_1457c0d6,radford2018improving}). 
Training large models requires larger datasets, longer training time and more computational resources, which leads to more energy consumption and carbon emission. 
For example, researchers in \cite{strubell2019energy} measured the energy consumption of  training large language models, and found that the estimated carbon footprint of training a single Transformer with neural architecture search emits approximately 300 tons of carbon dioxide~\cite{strubell2019energy}, which is of the order of 60 years of an average human being's carbon emission.

In the mean time, many works have been proposed to run deep models 
more efficiently. 
One type of works focuses on developing algorithms to 
obtain more efficient neural networks, such as pruning~\cite{hu2016network,molchanov2016pruning}, quantization\cite{banner2018post, fang2020post}, distillation~\cite{hinton2015distilling} and neural architecture search~\cite{zoph2016neural}. 
The other type of works aims at building more efficient platforms to deploy deep learning models, such as Tensorflow Lite, Pytorch Mobile and ONNX-Runtime. 
In this paper, we focus on benchmarking the first category of works aiming at making deep models more efficient, called Efficient Deep Learning \cite{menghani2021efficient}. 
Most of the existing efficient deep learning methods measure the efficiency of a model by its computational complexity, such as floating point operations (FLOPs) and number of parameters. 
In this paper, we benchmark efficient deep learning methods by directly evaluating their energy costs under various settings.

\begin{figure}[t]
    \centering 
    \includegraphics[width=0.95\linewidth]{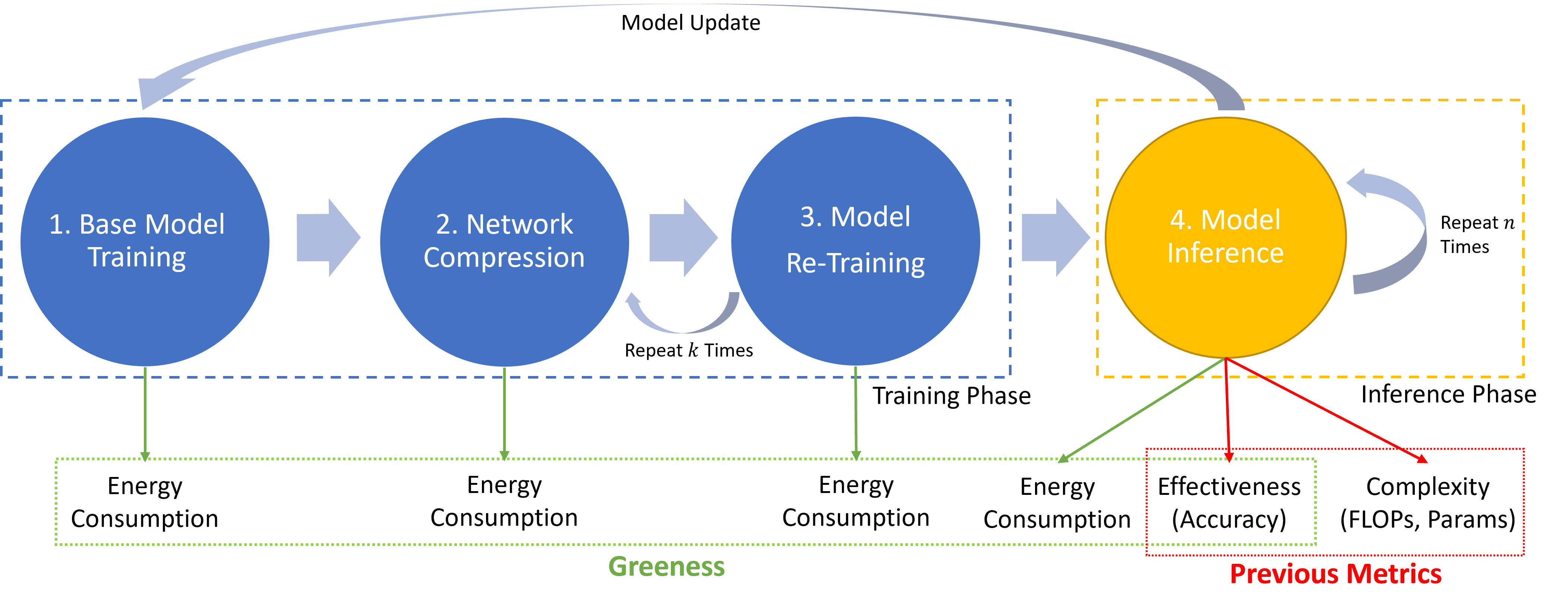}
    \caption{Life cycle of efficient deep learning. Most of the existing methods focus on the efficiency (in terms of model complexity) in the last stage, while we propose to directly evaluate the full-cycle energy consumption. The portion of training/inference phase energy consumption over the total energy consumption heavily corresponds to the model usage intensity. 
    }
    \vspace{-0.2cm}
    \label{fig:life_cycle}
\end{figure}

As shown in Figure \ref{fig:life_cycle}, we divide the entire life cycle of most existing efficient deep learning methods into four stages: 1) base model training, 2) network compression, 3) model re-train, and 4) compressed model inference, where the first 3 stages are the training phase and the last stage is the inference phase. Note that not all methods contain all four stages, as shown in Table \ref{tab:stages}.

In a standard AI system development cycle, four stages  usually form a closed loop. As shown in Figure \ref{fig:life_cycle}, when 
model updating is required, the life-cycle starts over from the beginning. 
As a result, the usage intensity of the compressed model becomes one of the key factors to the overall energy consumption. 
If the model needs to be frequently updated and only very few inferences can be performed in one life cycle, large portion of the total energy cost will come from training phase. Otherwise, if inference number per cycle is large, inference energy cost will dominate the overall energy consumption. 
As a result, we define the number of inferences conducted in one life-cycle as the \emph{model usage intensity} (MUI). 

Although many existing works are able to obtain lower energy consumption models at the inference stage, most of them neglect the energy cost of the first three stages. 
To fairly and comprehensively compare the energy cost of existing efficient deep learning methods, it is essential to consider the energy cost over the entire life cycle and under different MUIs. 
Note that in this paper we focus on the case where a new cycle always starts over from the first stage, other life-cycle scheme can also be evaluated with the same  methodology.
The main contributions of this paper are summarized as follows:
\begin{itemize}[noitemsep, leftmargin=*]
    \item \textbf{Full-Cycle Energy Consumption Metric: \green{Greeness}.} A new metric is proposed to consider the trade-offs among model effectiveness and energy consumption throughout the entire efficient deep  learning life cycle under different MUIs.
    \item \textbf{Energy Consumption Benchmark for Efficient Deep Learning.} Efficient deep learning baselines are comprehensively compared and analyzed based on the proposed metric in computer vision tasks, which could provide some insights for low-carbon computer vision in real applications.
    
\end{itemize}
\section{Related Work}
\begin{table}[htp]
\centering
\caption{Life-cycle stages in different types of efficient deep learning methods. }
\label{tab:stages}
\setlength{\tabcolsep}{1em} 
{\renewcommand{\arraystretch}{1.35}
\resizebox{\textwidth}{!}{%
\begin{tabular}{c|c|c|c|c}
\toprule
\multirow{2}{*}{Algorithm} & \multicolumn{3}{c|}{Model Training}                                               & \multirow{2}{*}{Model Inference} \\ \cline{2-4}
                           & Base Model Training       & Model Compression         & Model Re-training         &                                  \\ \hline
Traditional Deep Learning  & \cmark & \xmark     & \xmark & \cmark \\ \hline
Pruning                    & \cmark & \cmark & \cmark & \cmark        \\ \hline
Distillation               & \cmark & \xmark     & \cmark & \cmark        \\ \hline
Neural Architecture Search & \xmark     & \cmark & \cmark & \cmark        \\ \hline
Quantization (Post)        & \cmark & \cmark & \xmark & \cmark \\ \bottomrule
\end{tabular}%
}
}
\end{table}

\subsection{Efficient Deep Learning}
As mentioned above, we divide the life cycle of efficient deep learning into four stages. As shown in Table \ref{tab:stages}, different types of efficient deep learning methods contain different stages in their life cycles, and we introduce the details in this section. 

\textbf{Pruning.} Network pruning obtains smaller and faster models by reducing the number of parameters from a base model, whose training phase contains all three stages of the life-cycle. It first trains the base model, and then compresses the large model by removing the unimportant neurons in each layer. Finally, the pruned model is re-trained on the dataset to obtain the compressed model. 
The APoZ method~\cite{hu2016network} prunes the network based on the average percentage of zeros of a neuron from a set of validation examples. Li~\etal~\cite{DBLP:conf/iclr/0022KDSG17} proposed to prune filters from convolution layers based on filters' L1 norms. The FPGM method~\cite{he2019filter} is a data-independent method that prunes the filters based on their distances to the geometric mean of all filters. Two recent methods~\cite{molchanov2019importance,molchanov2016pruning} proposed to evaluate a filter's importance based on the error induced by removing it.

\textbf{Quantization.} Neural network quantization obtains smaller models by reducing the precision of the weights in the neural network. Post-training quantization methods~\cite{banner2018post, cai2020zeroq, choukroun2019low, fang2020post} contain two stages in the life-cycle, where it first trains a base model and then calibrates to compute the clipping ranges and the scaling factors of the quantization function with a small training set for model compression. The quantization-aware training methods~\cite{fan2020training, zhuang2018towards} contain three stages in the life-cycle, where  a pre-trained base model is quantized (compressed) and then finetuned using the training set. 

\textbf{Distillation.} Knowledge distillation~\cite{hinton2015distilling} obtains smaller and faster student model by learning from both the ground-truth labels and the outputs from a larger teacher model. Distillation contains two stages in the training phase of the life cycle: 1) training a large base model and 2) training a usually manually designed small student model under the supervision of ground-truth and teacher model in the re-training stage.  
Following the idea above, 
the PKT method~\cite{passalis2020probabilistic} minimizes the  KL divergence of the probability distributions between teacher model and student model. The CRD method~\cite{tian2019contrastive} uses a contrastive objective to transfer knowledge. 
Different from the above works, Komodakis~\etal~\cite{komodakis2017paying} proposed to distill attention maps from multiple network layers. 

\textbf{Neural Architecture Search.} We focus on the subset of Neural Architecture Search (NAS) methods that aims at obtaining more efficient neural networks in this paper. They automatically search a network architecture by jointly optimizing the model efficiency and accuracy. NAS has two stages in the training phase of life cycle. It does not need base model training, and directly searches a new network structure for re-training. One type of methods (e.g., EfficientNAS~\cite{pham2018efficient},  DARTS~\cite{liu2018darts}, ProxylessNAS~\cite{cai2018proxylessnas} and OFA~\cite{cai2019once}) first builds a super-net containing all the candidate architectures and searches for an optimal sub-model in one trial. In a way, this searching process is similar to model compression that compresses a super-net formed by a search space into a small network structure. 
Another family of methods like NASNet~\cite{zoph2016neural}, PNAS~\cite{liu2018progressive} and LargeEvo~\cite{real2017large} needs to train hundreds to thousands of candidate architectures (e.g., 12k~\cite{zoph2016neural}), consuming an excessively large amount of energy. Since multi-trial NAS methods cost about a hundred times more energy than one-trial methods~\cite{cai2019once} and bring no significant performance improvement, we focus on benchmarking one-trial NAS methods in this paper. 

\subsection{Green AI}
Schwartz~\etal~\cite{schwartz2020green} proposed a concept of Green AI, which refers to AI research that yields novel results while taking into account the computational cost, encouraging a reduction in resources spent. They mainly measure the greeness of neural networks by their number of float parameter operations, and neglect the energy consumption in the training phase. In this paper, we give a more comprehensive definition on greeness that considers the entire deep learning life-cycle under different model usage intensities. Patterson~\etal~\cite{patterson2021carbon} compared the carbon emission of training large NLP models on popular cloud servers. Strubell~\etal~\cite{strubell2019energy} also measured the training energy cost of existing NLP models and estimated the corresponding financial and environmental costs. Different from the above works, this paper targets at efficient deep learning in CNN models, and more importantly the proposed  \green{Greeness} metric and benchmarking method are orthogonal to the above works.

\section{\green{Greeness} in Efficient Deep Learning \label{sec:metric}}

In this section, we introduce a new metric --- \green{Greeness} to evaluate the efficient deep learning algorithms by directly measuring their energy consumption during the entire life cycle. 

\subsection{Key Factors of Greenness}
To achieve our goal, we consider the following key factors when designing the \green{Greeness} metric. 

\textbf{Train Energy Cost (TEC)}. TEC computes the overall energy consumption of an efficient deep learning algorithm throughout the entire training phase, including base model training, model compression and model re-training. 
For different algorithms, the specific composition of TEC may be different. In Table~\ref{tab:composition}, we summarize the TEC compositions for different types of algorithms. 

\textbf{Inference Energy Cost (IEC)}. IEC denotes the energy consumption of using the compressed model to perform inference for one time. 

\textbf{Model Usage Intensity (MUI)}. MUI is defined as the average number of inferences in each life-cycle. The importance of TEC and IEC varies based on the model usage intensity. If an AI system intensely use the model and the number of inferences is large in one life cycle, then a large proportion of the energy consumption comes from IEC, and vice versa. 

\textbf{Accuracy (Acc)}. Acc denotes the accuracy of the compressed model on a specific CV task. Efficient deep learning algorithms usually trade accuracy for efficiency and their accuracy degradation may vary significantly, so that we should also consider the accuracy of the compressed models.

\begin{table}[htp]
\centering
\caption{The full-cycle energy cost of different types of efficient deep learning methods. }
\label{tab:composition}
\setlength{\tabcolsep}{1em} 
{\renewcommand{\arraystretch}{1.5}
\resizebox{\textwidth}{!}{%
\begin{tabular}{c|c|c|c|c|c}
\toprule
\textbf{Algorithm} &
  \multicolumn{3}{c|}{\textbf{Train Energy Cost (TEC)}} &
  \multirow{5}{*}{\textbf{\begin{tabular}[c]{@{}c@{}}Inference Energy Cost\\ (IEC)\end{tabular}}} &
  \multirow{5}{*}{\textbf{\begin{tabular}[c]{@{}c@{}}Accuracy\\ (Acc.)\end{tabular}}} \\ \cline{1-4}
Pruning                    & \multirow{3}{*}{Train Original Model Cost} & Prune Cost        & \multirow{2}{*}{Finetune Cost} &  &  \\ \cline{1-1} \cline{3-3}
Quantization               &                                            & Quantization Cost &                                &  &  \\ \cline{1-1} \cline{3-4}
Distillation               &                                            & \multicolumn{2}{c|}{Distillation Cost}             &  &  \\ \cline{1-4}
Neural Architecture Search & Search Cost                                & \multicolumn{2}{c|}{Retrain Cost}                  &  &  \\ \bottomrule
\end{tabular}%
}
}
\end{table}

\subsection{Metric Design}

Given an efficient deep learning algorithm, let \texttt{Acc} denote the accuracy on a targeted task (between 0 and 1), \texttt{IEC} denotes the energy consumption for each inference time, \texttt{TEC} denotes the training energy cost of an efficient deep learning algorithm. We follow the idea of multiple objective evaluation (MOE)~\cite{DBLP:journals/access/BlankD20,velasquez2013analysis} to achieve the balance among multiple factors in the proposed metric \green{$\mathbb{G}$}. More specifically, we define the \green{Greeness} metric as follows:
\begin{equation}
    \label{eq:G}
    \green{$\mathbb{G}(\texttt{MUI})$} = \frac{\texttt{Acc}^{\tau}}{\texttt{MUI} * \texttt{IEC} + \texttt{TEC}}.
\end{equation}
\green{$\mathbb{G}(\texttt{MUI})$} is a trade-off between energy consumption and model performance over one entire life cycle, where the denominator measures the total energy consumption for one model life cycle and the numerator measures the model performance. $\tau$ is a hyper-parameter that indicates the tolerance to the accuracy loss brought by model compression. Higher $\tau$ value indicates that \G has a more strict requirement on the model accuracy  \texttt{Acc}. 

As shown in Equation~\ref{eq:G}, \green{$\mathbb{G}$} is a function of the model usage intensity, and efficient deep learning methods perform  differently under different model usage intensities. Intuitively, methods with higher $\mathtt{TEC}$ will perform better with high model usage intensity, (i.e., when $\texttt{MUI}$ is large), and vice versa. 

In the next section, we analyze and compare the \green{Greeness} of efficient deep learning methods under different model usage intensities.

\subsection{\green{Greeness} metric's relation with carbon emission}
In this paper, \green{Greeness} specifically refers to the carbon emission status of efficient deep learning methods, not carbon emission in a general sense. We argue that carbon emission under the scope of efficient deep learning is almost equivalent to energy consumption. This is because factors other than energy consumption (e.g., energy-mix) is independent to the methods, and can be kept fixed for a fair comparison. As stated in~\cite{patterson2021carbon}, CO2 emission is linearly proportional to energy consumption, and hence the \green{Greeness} in terms of energy consumption can be scaled to the \green{Greeness} in terms of CO2 emission with a constant value $C$, as derived in following equation.

\begin{equation}
\begin{aligned}
\mathbb{G}_{\texttt{CO2e}}&(\texttt{MUI})  := \frac{\texttt{Acc}^{\tau}}{\texttt{MUI} * \texttt{Inference}_{\texttt{CO2e}}+\texttt{Training}_{\texttt{CO2e}}} 
\\
& \texttt{Inference}_{\texttt{CO2e}} = C * \texttt{IEC}
\\
& \texttt{Training}_{\texttt{CO2e}} = C * \texttt{TEC}
\end{aligned}
\end{equation}

\section{\green{Greeness} Benchmark}
In this section, we present the details of the proposed \green{Greeness} benchmark for efficient deep learning of CNN models. Specifically, we investigate four types of efficient deep learning methods, namely pruning, quantization, knowledge distillation and neural architecture search.

\subsection{Experiment Configuration}

We first present the experiment configuration for obtaining the results in this section as follows. 

\textbf{Hardware and Platform} We run the training phase of the methods on Nvidia Tesla P100, and run the inference phase on Nvidia TITAN V to get a better hardware support on computations with \texttt{FP16} and \texttt{INT8} precision.  The influence of different GPU hardware on energy consumption will also be discussed in this section. PyTorch~\cite{paszke2019pytorch} is adopted as our model training framework. TensorRT~\footnote{\url{https://github.com/NVIDIA/TensorRT}} is adopted as the model deployment framework at inference time. 

\textbf{Energy Measurement} Since both the training and inference of the models are performed on GPU, IEC and TEC are measured by the energy consumption of GPU. We build a GPU energy tracer based on the \lstinline{nvidia-smi} interface, from which we get and parse the runtime information of the GPU. We add tracer functions at the corresponding locations for different algorithm implementations (e.g., train function, inference function, etc.). We ensure the tracer is completely added to the execution part of an algorithm to avoid excess errors as much as possible. Tracer function will open a separate thread to query runtime information per moment according to a predefined hyperparameter \textit{sampling frequency}. At the end of the main function, the tracer is killed and the total information collected by the tracer is recorded.

\textbf{Quantitative Scale} We choose W$\cdot$h as the unit for TEC and IEC. Note that other hardware components, e.g., CPU, memory, disk and cooling system, also consume energy, but since this work focuses on the deep learning scenarios where GPU takes up the majority of the total energy consumption~\cite{patterson2021carbon,strubell2019energy}, impacts of the other devices are neglected.

\textbf{Compared Methods} Greeness of the compared methods are evaluated by classification task \textit{CIFAR100}. 
The hyper-parameter settings of the compared methods are identical to the original papers to ensure fair comparison for both intra-type and cross-type algorithms. 
Since no existing repository has implemented all types of algorithms, there are naturally some inevitable errors in cross-type algorithms comparisons. 
We ensure the fairness of the comparison through a reasonable experimental setup. Below is the detailed introduction of the compared methods. Note that in this section, we only reported the selected results due to space limitation. Complete experiment results can be found in the Appendix. 

\begin{itemize}[noitemsep, leftmargin=*]
\item \textbf{Baselines} Neural Network Intelligence (NNI)\footnote{\url{https://github.com/microsoft/nni}} framework is used to train selected baseline models including VGG  variants~\cite{DBLP:journals/corr/SimonyanZ14a} and ResNet variants ~\cite{DBLP:conf/eccv/HeZRS16}. 
Note that to adapt the image resolution of CIFAR100, following the implementation in NNI, some networks like VGG and ResNet18 are adjusted by removing some of the downsampling operations. 

\item \textbf{Pruning} The pruning methods are evaluated based on NNI. Following the default practice of NNI, we consider two types of pruners: one-shot pruners (L1 Filter~\cite{DBLP:conf/iclr/0022KDSG17}, L2 Filter~\cite{DBLP:conf/iclr/0022KDSG17}, FPGM~\cite{he2019filter}) and iterative pruners (APoZ~\cite{hu2016network},  TaylorFO~\cite{molchanov2019importance}, Activation Mean~\cite{DBLP:conf/iclr/MolchanovTKAK17}). For both types of methods, the base models (including ResNet 18/34/50, VGG 16/19) are firstly trained for 160 epochs. The base models are then pruned and finally finetuned with aformentioned pruners. The total finetune epoch for iterative pruners is set to 160.

\item \textbf{Quantization } Here we choose a simple yet effective low-energy-cost post training quantization method as baseline.  Symmetric uniform quantizations implemented in TensorRT with both FP16 and INT8 precision are evaluated. 

\item  \textbf{Distillation} The distillation methods are evaluated based on 
{RepDistiller}\footnote{\url{https://github.com/HobbitLong/RepDistiller}}. In total, 13 state-of-the-art algorithms are benchmarked, including KD~\cite{hinton2015distilling}, FitNet~\cite{romero2014fitnets}, AT~\cite{komodakis2017paying}, SP~\cite{tung2019similarity}, CC~\cite{peng2019correlation}, VID~\cite{ahn2019variational}, RKD~\cite{park2019relational}, PKT~\cite{passalis2020probabilistic}, AB~\cite{heo2019knowledge}, FT~\cite{kim2018paraphrasing}, FSP~\cite{yim2017gift}, NST~\cite{huang2017like} and CRD~\cite{tian2019contrastive}. They are all tested on different architectural types. According to the default configuration, the teacher models are trained for 240 epochs, and then distillated for 240 epochs to obtain student models. The accuracy of the distillation methods is cited from \cite{tian2019contrastive}. 

\item  \textbf{Neural Architecture Search} The Neural Architecture Search (NAS) evaluation is also based on NNI~\cite{zhang2020retiarii}. Since multi-trial NAS methods are too costly without significant performance improvement~\cite{cai2019once} compared to our one-trail alternatives, we focus on benchmarking one-trial NAS methods. In the experiments, two of the most popular one-shot NAS methods are reported, namely DARTS~\cite{liu2018darts} and ProxylessNAS~\cite{cai2018proxylessnas}. 

\end{itemize}

\begin{table}
\centering
\caption{Selected results for different types of efficient deep learning methods in terms of TEC, IEC, ACC and \green{$\mathbb{G}$}. Accuracy is measured on \textit{CIFAR100} and accurarcy tolerance $\tau$ is set to 2. \green{$\mathbb{G}$} (500M) and \green{$\mathbb{G}$} (1B) indicate the greeness value when MUI = $5*10^{8}$ and $1*10^{9}$.}
\label{tab:perf}
\resizebox{\linewidth}{!}{%
\begin{tabular}{c|c|c|c|c|c|c|c|c}
\toprule
\multirow{4}{*}{Baseline} & \multicolumn{3}{c|}{Model} & TEC & Acc. & IEC & \green{$\mathbb{G}$} (500M) & \green{$\mathbb{G}$} (1B) \\ 
\cline{2-9}
 & \multicolumn{3}{c|}{ResNet18} & 276.27 & 76.56 & 1.13E-06 & 6.95 & 4.16 \\ 
\cline{2-9}
 & \multicolumn{3}{c|}{VGG16} & 138.59 & 73.37 & 8.93E-07 & 9.20 & 5.22 \\ 
\cline{2-9}
 & \multicolumn{3}{c|}{VGG19} & 179.10 & 72.51 & 1.01E-06 & 7.68 & 4.42 \\ 
\hline
\multirow{5}{*}{Distillation} & Teacher Model & Student Model & Method & TEC & Acc. & IEC & \green{$\mathbb{G}$} (500M) & \green{$\mathbb{G}$} (1B) \\ 
\cline{2-9}
 & \multirow{2}{*}{VGG 13} & \multirow{2}{*}{VGG 8} & CRD & 287.45 & 73.94 & 5.88E-07 & 9.41 & 6.25 \\ 
\cline{4-9}
 &  &  & KD & 181.61 & 72.98 & 5.39E-07 & 11.80 & 7.39 \\ 
\cline{2-9}
 & \multirow{2}{*}{ResNet 56} & \multirow{2}{*}{ResNet 20} & CRD & 244.74 & 71.16 & 4.78E-07 & 10.47 & 7.01 \\ 
\cline{4-9}
 &  &  & KD & 259.58 & 70.66 & 4.68E-07 & 10.12 & 6.87 \\ 
\hline
\multirow{3}{*}{Quantization} & \multicolumn{2}{c|}{Model} & Method & TEC & Acc. & IEC & \green{$\mathbb{G}$} (500M) & \green{$\mathbb{G}$} (1B) \\ 
\cline{2-9}
 & \multicolumn{2}{c|}{\multirow{2}{*}{VGG16}} & FP16 & 138.59 & 73.23 & 5.86E-07 & 12.43 & 7.40 \\ 
\cline{4-9}
 & \multicolumn{2}{c|}{} & INT8 & 138.59 & 73.22 & 5.72E-07 & 12.63 & 7.55 \\ 
\hline
\multirow{7}{*}{Pruning} & \multicolumn{2}{c|}{Model} & Method & TEC & Acc. & IEC & \green{$\mathbb{G}$} (500M) & \green{$\mathbb{G}$} (1B) \\ 
\cline{2-9}
 & \multicolumn{2}{c|}{\multirow{6}{*}{VGG16}} & APoZ Pruner & 154.27 & 70.59 & 5.61E-07 & 11.46 & 6.96 \\ 
\cline{4-9}
 & \multicolumn{2}{c|}{} & FPGM Pruner & 155.83 & 70.46 & 5.91E-07 & 11.00 & 6.64 \\ 
\cline{4-9}
\cline{4-9}
 & \multicolumn{2}{c|}{} & L2 Filter Pruner & 158.36 & 71.09 & 5.67E-07 & 11.44 & 6.97 \\ 
\cline{4-9}
 & \multicolumn{2}{c|}{} & TaylorFO Pruner & 146.70 & 70.69 & 5.65E-07 & 11.64 & 7.02 \\ 
\cline{4-9}
\hline
\multirow{5}{*}{NAS} & \multicolumn{2}{c|}{Search Space} & Method & TEC & Acc. & IEC & \green{$\mathbb{G}$} (500M) & \green{$\mathbb{G}$} (1B) \\ 
\cline{2-9}
 & \multicolumn{2}{c|}{PyramidNet with modifications (CIFAR100)} & ProxylessNAS & 280.78 & 77.61 & 6.48E-07 & 9.96 & 6.48 \\ 
\cline{2-9}
 & \multicolumn{2}{c|}{PyramidNet with modifications (CIFAR10)} & ProxylessNAS & 262.30 & 76.48 & 6.05E-07 & 10.35 & 6.74 \\ 
\cline{2-9}
 & \multicolumn{2}{c|}{Default CNN search space (CIFAR100)} & Darts & 4352.13 & 76.87 & 2.34E-06 & 1.07 & 0.88 \\ 
\cline{2-9}
 & \multicolumn{2}{c|}{Default CNN search space (CIFAR10)} & Darts & 3848.50 & 77.04 & 2.45E-06 & 1.17 & 0.94 \\
\bottomrule
\end{tabular}
}
\end{table}

\subsection{Analysis and Insights}
In this section, we compare four types of efficient deep learning methods, namely pruning, quantization, distillation and NAS. We mainly record each algorithm's training energy cost (TEC), the inference energy cost (IEC), accuracy, and some other GPU related information. In the following, we will present more analysis about the impacts of different factors on \green{Greeness}.

\subsubsection{Comparison across different types of methods}
We compare \green{Greeness} of different types of efficient deep learning methods under different MUIs, and select one representative method for each type. 
The methods are selected based on two criteria: 1) greeness comparison among the same type of methods, and 2) the base models are similar. 
As a result, five methods are selected, namely vgg16 as baseline, distillation with resnet13 as teacher and resnet8 as student, quantization on vgg16, pruning on vgg16 and Proxyless NAS on \textit{CIFAR100}. 

\begin{figure}[t!]
    \centering
    \resizebox{\textwidth}{!}{\includegraphics{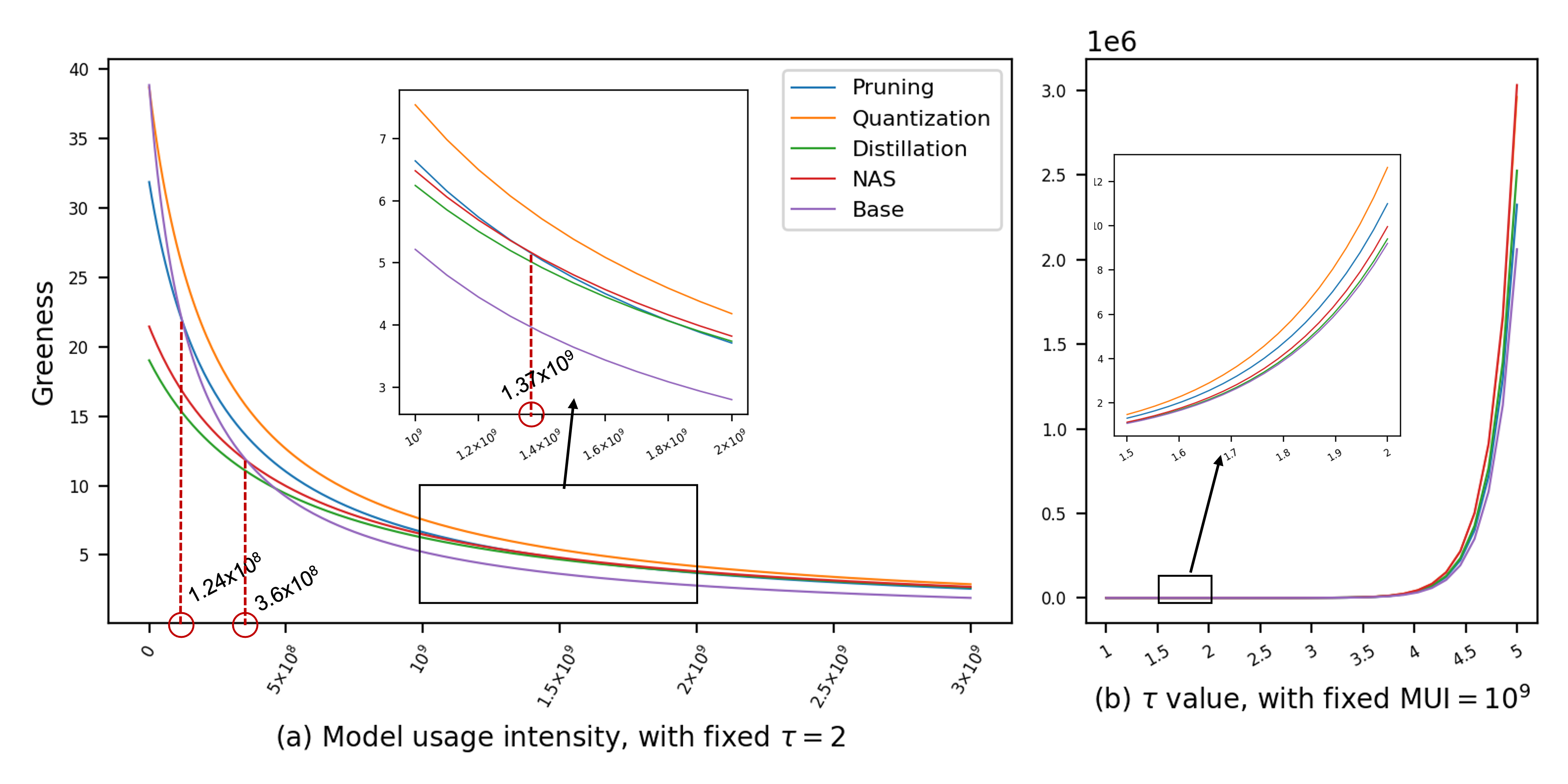}}
    \caption{With several selected algorithms of similar base models, we evaluate the curve of \green{Greeness} as $n$ increases. The selected ones are: (1) APoZ pruner on VGG16. (2) INT8 quantization on VGG16. (3) CRD distillati on from VGG13 to VGG8. (4) Proxyless NAS. (5) Basic VGG16 training. X-axis in (a) indicates MUI with $\tau$ is set to 2, and X-axis in (b) indicates $\tau$ value with MUI is set to $10^9$.}
    \label{fig:greeness_across_algos}
\end{figure}

As shown in Figure \ref{fig:greeness_across_algos} a), \green{Greeness} varies significantly under different MUIs, and different types of methods prevail within different MUI regions. To achieve the best carbon efficiency, we should carefully choose the suitable types of efficient deep learning methods for different application scenarios (high or low model usage intensity). 

From Figure \ref{fig:greeness_across_algos}, for each type of methods, we summarize the following guidelines on the suitable  scenarios for the methods to prevail. 

\begin{itemize}[noitemsep, leftmargin=*]
    \item \textbf{Low Model Usage}. 
    Pruning achieves higher greeness score when \texttt{MUI} is small, due to low TEC (shown in Table \ref{tab:perf}). It outperforms both NAS and Distillation when MUI is less than 1.37 billion. This shows in the scenarios where the model requires constant update (e.g., new project with relative flexible requirements), pruning is more suitable compared to high TEC methods like NAS and distillation. 
    Furthermore, when MUI is very low, except for quantization, all efficient learning methods show no improvement in terms of energy consumption. 
    Only training the base models outperforms all three types of methods expect quantization when MUI is less than 124 million.
    \item \textbf{High Model Usage}. As shown in Table \ref{tab:perf}, NAS achieves lower IEC and accuracy at the cost of high TEC. As a result, they achieve higher greeness when $\texttt{MUI}$ is large. 
    NAS begins to outperform the other methods not until the MUI increases to 1.37 billion. Similarly, distillation also achieves better greeness score under higher MUI. This shows that NAS and Distillaton are  more suitable in the high model usage scenarios, where the model is relatively stable and does not need to be updated often. 
    
    \item \textbf{Quantization}. Table \ref{tab:perf} shows that using post training quantization methods have almost no extra efficient learning cost compared to normal deep learning. It effectively reduces IEC with almost no accuracy loss. Hence, quantization constantly achieves high greeness score throughout different MUIs. It shows that quantization is a very general and effective method that should always be applied regardless of the application scenarios. However, quantization requires specific hardware support on different computation precision. 
\end{itemize}




In Figure~\ref{fig:greeness_across_algos} (b), we demonstrate the relationship between Greeness and $\tau$ value. The $\tau$ value reflects, to some extent, the importance attached to accuracy in the calculation of Greeness, and a larger $\tau$ means a larger share in the Greeness metric, corresponding to a scenario with a higher requirement for model accuracy. The figure shows that NAS algorithm has higher greeness score when $\tau$ is small and pruning algorithm has higher greeness score when $\tau$ is large.

\subsubsection{Comparison within the same type of methods}

In this section, we present the performance comparison of different efficient deep learning methods within the same type.

\textbf{Distillation.} The Greeness scores of distillation methods are affected by three entangled factors, namely the student model, teacher model and accuracy. Larger teacher model has higher TEC but is more likey to achieve higher accuracy.  Smaller student model leads to lower IEC but lower accuracy score. 

\begin{figure}[htb]
    \centering
    \resizebox{\textwidth}{!}{\includegraphics{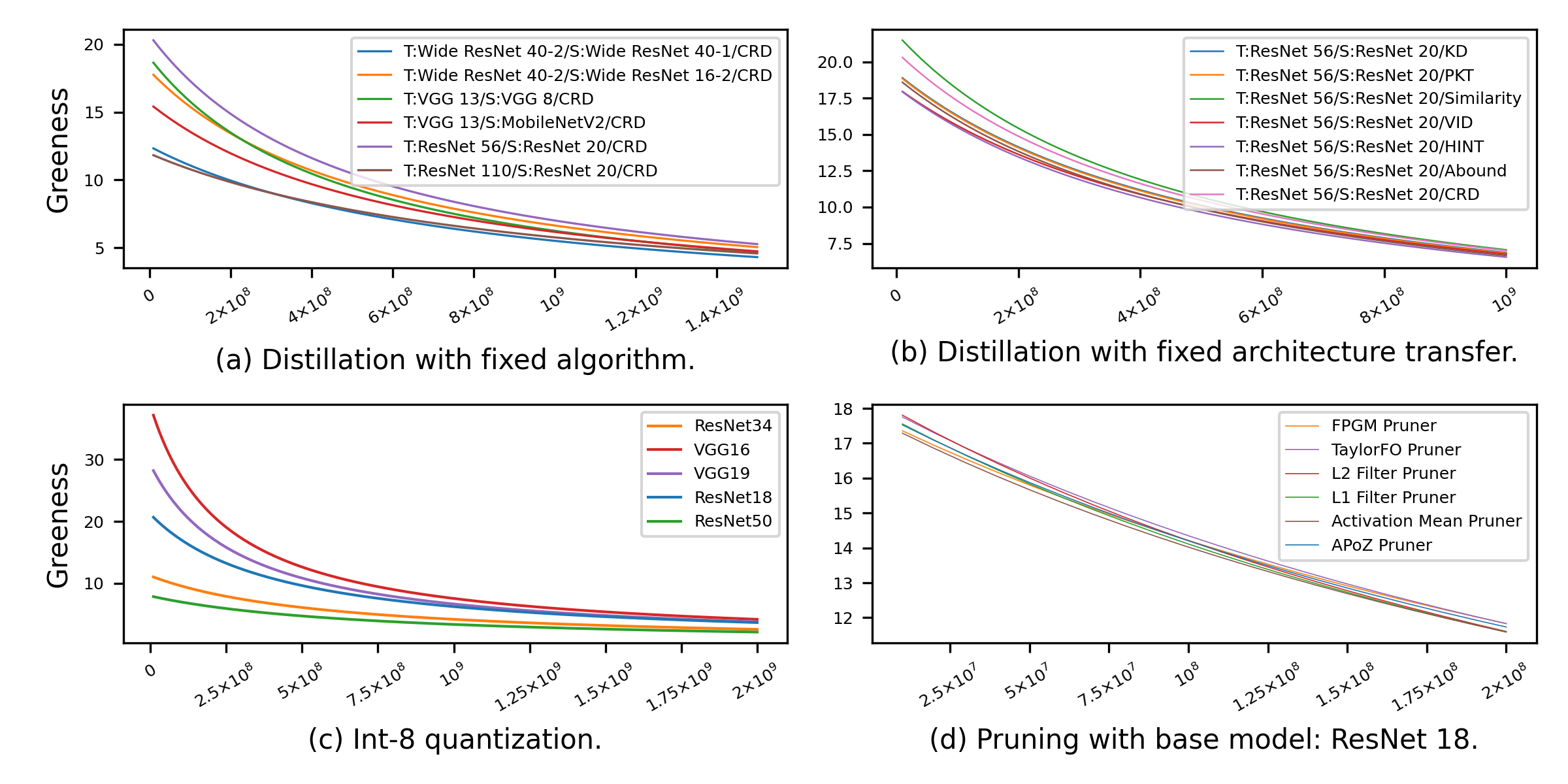}}
    \caption{Greeness (Y-axis) curve with different MUI (X-axis), the accuracy tolerance value $\tau$ is set to 2. (a) CRD~\cite{tian2019contrastive} with various types of teacher and student models. (b) ResNet 56 (20) as teacher (student) model, with various algorithms. (c) Int-8 quantization with various models. (d) ResNet 18 as base model with various pruning algorithms.}
    \vspace{-0.1cm}
    \label{fig:dist_prun}
\end{figure}

Figure \ref{fig:dist_prun} a) compares the CRD's Greeness score under different settings of teacher/student models. Figure \ref{fig:dist_prun} b) compares different distillation methods with the same teacher/student models. We observe that Similarity (SP)~\cite{tung2019similarity} and CRD~\cite{tian2019contrastive} have higher Greeness scores than the rest algorithms in our selected model architecture, but this advantage diminishes when MUI is larger than around $10^{9}$.

\textbf{Pruning.} Figure \ref{fig:dist_prun} a) shows the Greeness scores of applying different pruning methods on ResNet 18. We observe TaylorFO and APoZ pruner achieve highest Greeness when MUI is small, and FPGM Prunner outperforms when MUI is large. This is mainly because FPGM has larger TEC than APoZ and TaylorFO, but is able to obtain model with much lower IEC, as shown in Table \ref{tab:perf}. 

\textbf{Quantization.} We apply the same quantization methods on different neural networks and observe that smaller model seem to gain better greeness score when MUI is relatively small due to smaller TEC and larger tolerace on accuracy. 

\subsection{Sensitivity Analysis on Energy Consumption}
In this section, we report the results on  other factors that may affect the energy consumption beyond the efficient deep learning methods. We put more on supplementary materials.





\begin{figure}[!tp]
    \centering
    \resizebox{0.98\textwidth}{!}{\includegraphics{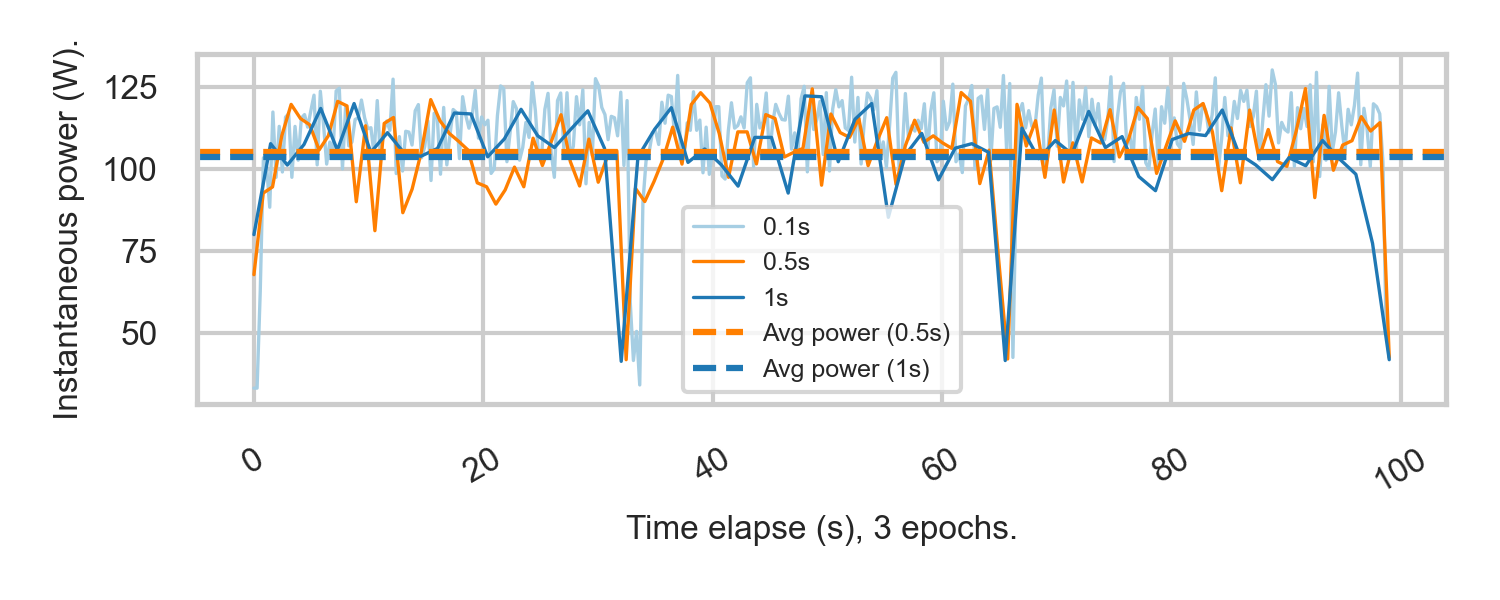}}
    \caption{To verify the functionality of our GPU Tracer and select a good sampling frequency. We record the instantaneous (runtime) power at different sampling frequencies during 3 epochs training. The trough (bottom) of instantaneous power indicates data loading between epochs (this part will also be ignored in the result of model energy consumption).}
    \label{fig:sf}
\end{figure}

\textbf{The influence of sampling frequency.}
We sample the GPU information once at each moment, and a higher sampling frequency corresponds to richer collection of information over time. In our experiments, we set the sampling frequency to 0.1s. Figure~\ref{fig:sf}  demonstrates how the instantaneous power varies with different sampling frequencies (e.g. \textit{sampling frequency} = 0.1s, 0.5s, 1s). The horizontal lines indicate the average power at different sampling frequencies throughout 3 epochs, in which the observed average power variations are negligible.


\textbf{The influence of GPU types.} Table \ref{tab:gpu_type} reports the energy cost of knowledge distillation algorithms across different GPU types. We observe that when running the same algorithm, newer GPU (e.g., V100) indeed have much shorter execution time and comparable average power, hence resulting in lower energy consumption.

\begin{figure}[tp]
    \centering 
    \resizebox{\textwidth}{!}{\includegraphics{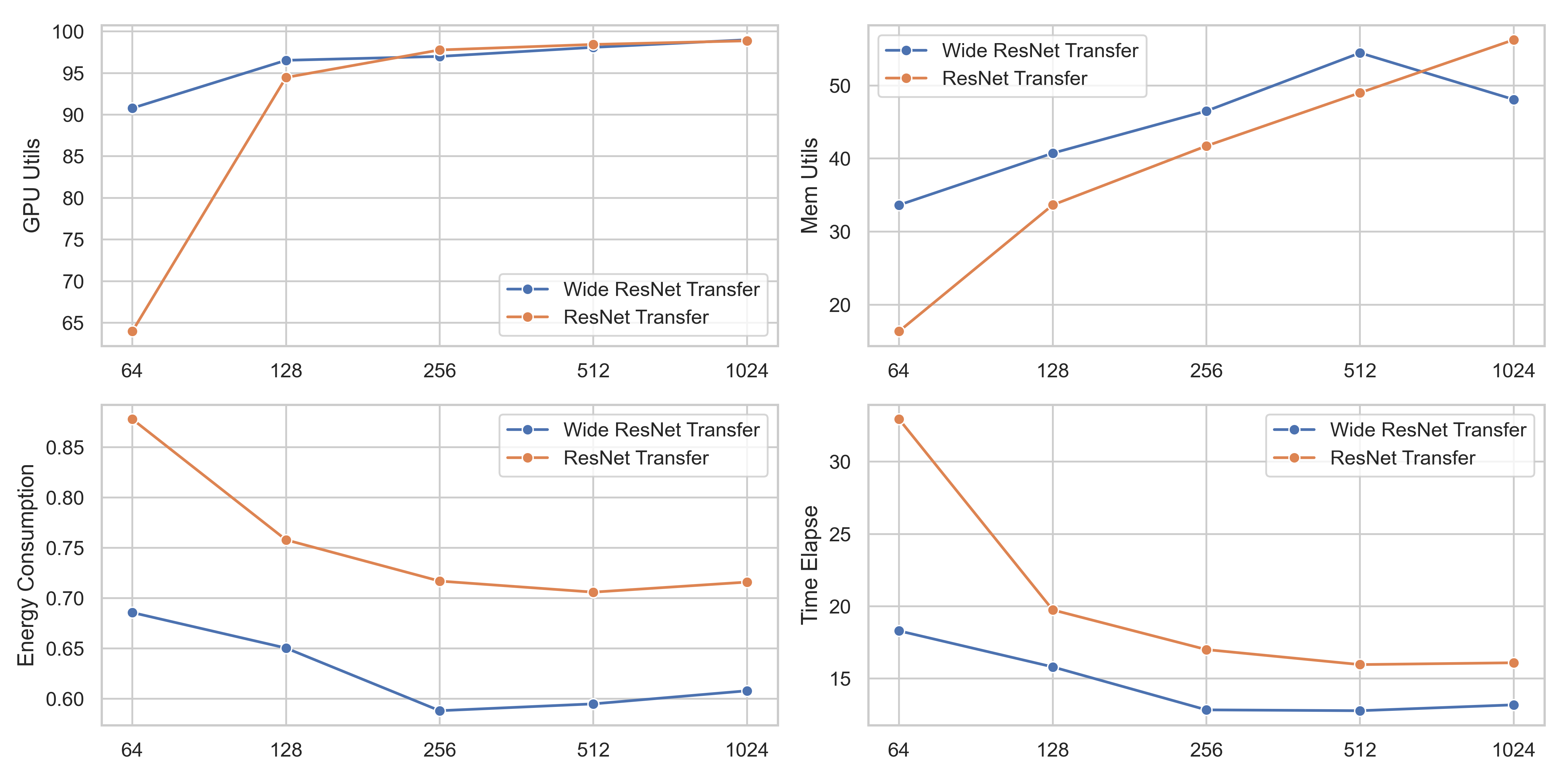}}
    \caption{GPU usage sensitivity over batch size. Results of (1) KD with teacher model Wide ResNet 40-2 to student model ResNet 16-2 (2) KD with teacher model Resnet 110 to student model ResNet 32 are reported. }. 
    \label{fig:batch_size}
\end{figure}

\textbf{The influence of batch size in different implementations.}
The influence of batch size on energy consumption is complex. 
On one hand, larger batch size accelerates the convergence which helps reduce GPU execution time. On the other hand, larger batch size increases the GPU utility rate and memory usage, which increases power consumption  during algorithm execution. 
In Figure~\ref{fig:batch_size}, with other settings fixed, we compare how the GPU information changes for different batch sizes. As can be seen, increasing batch size reduces the algorithm execution time and minimizes the total energy consumption when batch size is no larger than 256. After that, since the GPU already runs at the full capacity, increasing batch size does not help to reduce total energy consumption. 

\begin{table}[!tp]
\centering
\caption{GPU usage over different GPU types. Results of KD with Wide ResNet 40-2 as teacher and ResNet 16-2 as student are reported. }
\label{tab:gpu_type}
\resizebox{\linewidth}{!}{%
\begin{tabular}{c|c|c|c|c|c} 
\toprule
GPU Type & GPU Utils (\%) & Mem Utils (\%) & Avg Power (W) & Time Elapse (S) & Energy Consumption (W$\cdot$H) \\ 
\hline
NVIDIA TESLA P100 & 71.45 & 26.29 & 133.76 & 38.83 & 0.80 \\ 
\hline
NVIDIA TESLA P40 & 77.51 & 49.86 & 134.60 & 30.97 & 0.90 \\
\hline
NVIDIA TESLA V100 16GB & 35.00 & 19.00 & 171.88 & 14.61 & 0.64 \\
\bottomrule
\end{tabular}
}
\end{table}

\subsection{Limitations of our work}
The limitations of our works are stated in different parts of the paper. We stated the limitation on reporting the greenness on limited GPU types in section 4.1 and 4.3, the limitation on reporting on limited hyper-parameter settings (e.g., batch size) in section 4.3, and the limitation on not considering the energy consumption on running efficient deep learning on complex distributed system is stated in section 4.1.

\section{Conclusion}
The energy consumption and carbon emission of deep learning models have been increasing dramatically over the years. 
Although efficient deep learning techniques are able to obtain models with fewer parameters and smaller latency --- leading to low inference energy consumption, 
most of them neglect the notable energy consumption throughout the entire life cycle of the algorithm. 
In this paper, we present the first large-scale energy consumption benchmark for efficient deep learning in computer vision models, where a new metric is proposed to explicitly consider full-cycle energy consumption under different model usage intensity. 
The benchmark 
can provide insights for low carbon emission when selecting efficient deep learning algorithms in computer vision tasks under different model usage intensities. 

{\small
\bibliographystyle{plain}
\bibliography{main}
}
\section*{Checklist}

\begin{enumerate}

\item For all authors...
\begin{enumerate}
  \item Do the main claims made in the abstract and introduction accurately reflect the paper's contributions and scope?
    \answerYes{}{}
  \item Did you describe the limitations of your work?
    \answerYes{}
  \item Did you discuss any potential negative societal impacts of your work?
    \answerYes{}
  \item Have you read the ethics review guidelines and ensured that your paper conforms to them?
    \answerYes{}
\end{enumerate}

\item If you are including theoretical results...
\begin{enumerate}
  \item Did you state the full set of assumptions of all theoretical results?
    \answerNA{}
	\item Did you include complete proofs of all theoretical results?
    \answerNA{}
\end{enumerate}

\item If you ran experiments (e.g. for benchmarks)...
\begin{enumerate}
  \item Did you include the code, data, and instructions needed to reproduce the main experimental results (either in the supplemental material or as a URL)?
    \answerYes{}
  \item Did you specify all the training details (e.g., data splits, hyperparameters, how they were chosen)?
    \answerYes{}
	\item Did you report error bars (e.g., with respect to the random seed after running experiments multiple times)?
    \answerYes{}
	\item Did you include the total amount of compute and the type of resources used (e.g., type of GPUs, internal cluster, or cloud provider)?
    \answerYes{}
\end{enumerate}

\item If you are using existing assets (e.g., code, data, models) or curating/releasing new assets...
\begin{enumerate}
  \item If your work uses existing assets, did you cite the creators?
    \answerYes{}
  \item Did you mention the license of the assets?
    \answerYes{}
  \item Did you include any new assets either in the supplemental material or as a URL?
    \answerYes{}
  \item Did you discuss whether and how consent was obtained from people whose data you're using/curating?
    \answerYes{}
  \item Did you discuss whether the data you are using/curating contains personally identifiable information or offensive content?
    \answerYes{}
\end{enumerate}

\item If you used crowdsourcing or conducted research with human subjects...
\begin{enumerate}
  \item Did you include the full text of instructions given to participants and screenshots, if applicable?
    \answerNA{}
  \item Did you describe any potential participant risks, with links to Institutional Review Board (IRB) approvals, if applicable?
    \answerNA{}
  \item Did you include the estimated hourly wage paid to participants and the total amount spent on participant compensation?
    \answerNA{}
\end{enumerate}

\end{enumerate}

\include{supp}

\end{document}

%% file: supp.tex
\newpage
\appendix
\section{More results with different $\tau$ value}

In the main paper, we mainly provide the experimental results with $\tau$ value is set to 2. In fact, $\tau$ can be taken in different values according to different application scenarios, and here we show the results for other $\tau$ value (e.g. 5 and 10).
\begin{table}[!h]
\centering
\caption{Selected results for different types of efficient deep learning methods in terms of TEC, IEC, ACC and \green{$\mathbb{G}$}. Accuracy is measured on \textit{CIFAR100} and accurarcy tolerance $\tau$ is set to 5 and 10. \green{$\mathbb{G}$} (500M, 5) indicates MUI = $5*10^{8}$, $\tau$=5 and \green{$\mathbb{G}$} (500M, 10) indicates MUI = $5*10^{8}$, $\tau$=10.}
\label{tab:perf_2}
\resizebox{\textwidth}{!}{%
\begin{tabular}{c|c|c|c|c|c|c|c|c}
\toprule
\multirow{4}{*}{\textbf{Baseline}}     & \multicolumn{3}{c|}{\textbf{Model}}                                                      & \textbf{Train Energy   Cost} & \textbf{Accuracy} & \textbf{Inference   Energy Cost} & \textbf{\green{$\mathbb{G}$}   (500M, 5)} & \textbf{\green{$\mathbb{G}$}   (500M, 10)} \\ \cline{2-9} 
                                       & \multicolumn{3}{c|}{ResNet18}                                                            & 276.27                       & 76.56             & 1.134E-06                        & 3.12E+06                                  & 8.20E+15                                   \\ \cline{2-9} 
                                       & \multicolumn{3}{c|}{VGG16}                                                               & 138.59                       & 73.37             & 8.931E-07                        & 3.63E+06                                  & 7.73E+15                                   \\ \cline{2-9} 
                                       & \multicolumn{3}{c|}{VGG19}                                                               & 179.10                       & 72.51             & 1.011E-06                        & 2.93E+06                                  & 5.87E+15                                   \\ \hline
\multirow{5}{*}{\textbf{Distillation}} & \textbf{Teacher Model}          & \textbf{Student Model}        & \textbf{Method}        & \textbf{Train Energy Cost}   & \textbf{Accuracy} & \textbf{Inference Energy Cost}   & \textbf{\green{$\mathbb{G}$}   (500M, 5)} & \textbf{\green{$\mathbb{G}$}   (500M, 10)} \\ \cline{2-9} 
                                       & \multirow{2}{*}{VGG   13}       & \multirow{2}{*}{VGG 8}        & CRD                    & 287.45                       & 73.94             & 5.876E-07                        & 3.80E+06                                  & 8.40E+15                                   \\ \cline{4-9} 
                                       &                                 &                               & KD                     & 181.61                       & 72.98             & 5.393E-07                        & 4.59E+06                                  & 9.50E+15                                   \\ \cline{2-9} 
                                       & \multirow{2}{*}{ResNet   56}    & \multirow{2}{*}{ResNet 20}    & CRD                    & 244.74                       & 71.16             & 4.78E-07                         & 3.77E+06                                  & 6.88E+15                                   \\ \cline{4-9} 
                                       &                                 &                               & KD                     & 259.58                       & 70.66             & 4.68E-07                         & 3.57E+06                                  & 6.29E+15                                   \\ \hline
\multirow{3}{*}{\textbf{Quantization}} & \multicolumn{2}{c|}{\textbf{Model}}                             & \textbf{Method}        & \textbf{Train Energy Cost}   & \textbf{Accuracy} & \textbf{Inference Energy Cost}   & \textbf{\green{$\mathbb{G}$}   (500M, 5)} & \textbf{\green{$\mathbb{G}$}   (500M, 10)} \\ \cline{2-9} 
                                       & \multicolumn{2}{c|}{\multirow{2}{*}{VGG16}}                     & FP16                   & 138.59                       & 73.23             & 5.85794E-07                      & 4.88E+06                                  & 1.03E+16                                   \\ \cline{4-9} 
                                       & \multicolumn{2}{c|}{}                                           & INT8                   & 138.59                       & 73.22             & 5.717E-07                        & 4.96E+06                                  & 1.04E+16                                   \\ \hline
\multirow{7}{*}{\textbf{Pruning}}      & \multicolumn{2}{c|}{\textbf{Model}}                             & \textbf{Method}        & \textbf{Train Energy Cost}   & \textbf{Accuracy} & \textbf{Inference Energy Cost}   & \textbf{\green{$\mathbb{G}$}   (500M, 5)} & \textbf{\green{$\mathbb{G}$}   (500M, 10)} \\ \cline{2-9} 
                                       & \multicolumn{2}{c|}{\multirow{6}{*}{VGG16}}                     & APoZ Pruner            & 154.27                       & 70.59             & 5.612E-07                        & 4.03E+06                                  & 7.06E+15                                   \\ \cline{4-9} 
                                       & \multicolumn{2}{c|}{}                                           & FPGM Pruner            & 155.83                       & 70.46             & 5.914E-07                        & 3.85E+06                                  & 6.68E+15                                   \\ \cline{4-9} 
                                       & \multicolumn{2}{c|}{}                                           & L1 Filter Pruner       & 155.77                       & 71.88             & 5.634E-07                        & 4.39E+06                                  & 8.42E+15                                   \\ \cline{4-9} 
                                       & \multicolumn{2}{c|}{}                                           & L2 Filter Pruner       & 158.36                       & 71.09             & 5.670E-07                        & 4.11E+06                                  & 7.46E+15                                   \\ \cline{4-9} 
                                       & \multicolumn{2}{c|}{}                                           & TaylorFO Pruner        & 146.70                       & 70.69             & 5.650E-07                        & 4.11E+06                                  & 7.26E+15                                   \\ \cline{4-9} 
                                       & \multicolumn{2}{c|}{}                                           & Activation Mean Pruner & 155.71                       & 70.76             & 5.621E-07                        & 4.06E+06                                  & 7.21E+15                                   \\ \hline
\multirow{5}{*}{\textbf{NAS}}          & \multicolumn{2}{c|}{\textbf{Search   Space}}                    & \textbf{Method}        & \textbf{Train   Energy Cost} & \textbf{Accuracy} & \textbf{Inference   Energy Cost} & \textbf{\green{$\mathbb{G}$}   (500M, 5)} & \textbf{\green{$\mathbb{G}$}   (500M, 10)} \\ \cline{2-9} 
                                       & \multicolumn{2}{c|}{PyramidNet   with modifications (CIFAR100)} & Proxyless   Nas        & 280.78                       & 77.61             & 6.481E-07                        & 4.66E+06                                  & 1.31E+16                                   \\ \cline{2-9} 
                                       & \multicolumn{2}{c|}{PyramidNet   with modifications (CIFAR10)}  & Proxyless   Nas        & 262.30                       & 76.48             & 6.052E-07                        & 4.63E+06                                  & 1.21E+16                                   \\ \cline{2-9} 
                                       & \multicolumn{2}{c|}{Default   CNN search space (CIFAR100)}      & Darts                  & 4352.13                      & 76.87             & 2.339E-06                        & 4.86E+05                                  & 1.30E+15                                   \\ \cline{2-9} 
                                       & \multicolumn{2}{c|}{Default   CNN search space (CIFAR10)}       & Darts                  & 3848.50                      & 77.04             & 2.445E-06                        & 5.35E+05                                  & 1.45E+15                                   \\ \bottomrule
\end{tabular}%
}
\end{table}

\begin{figure}[h]
    \centering
    \resizebox{\textwidth}{!}{\includegraphics{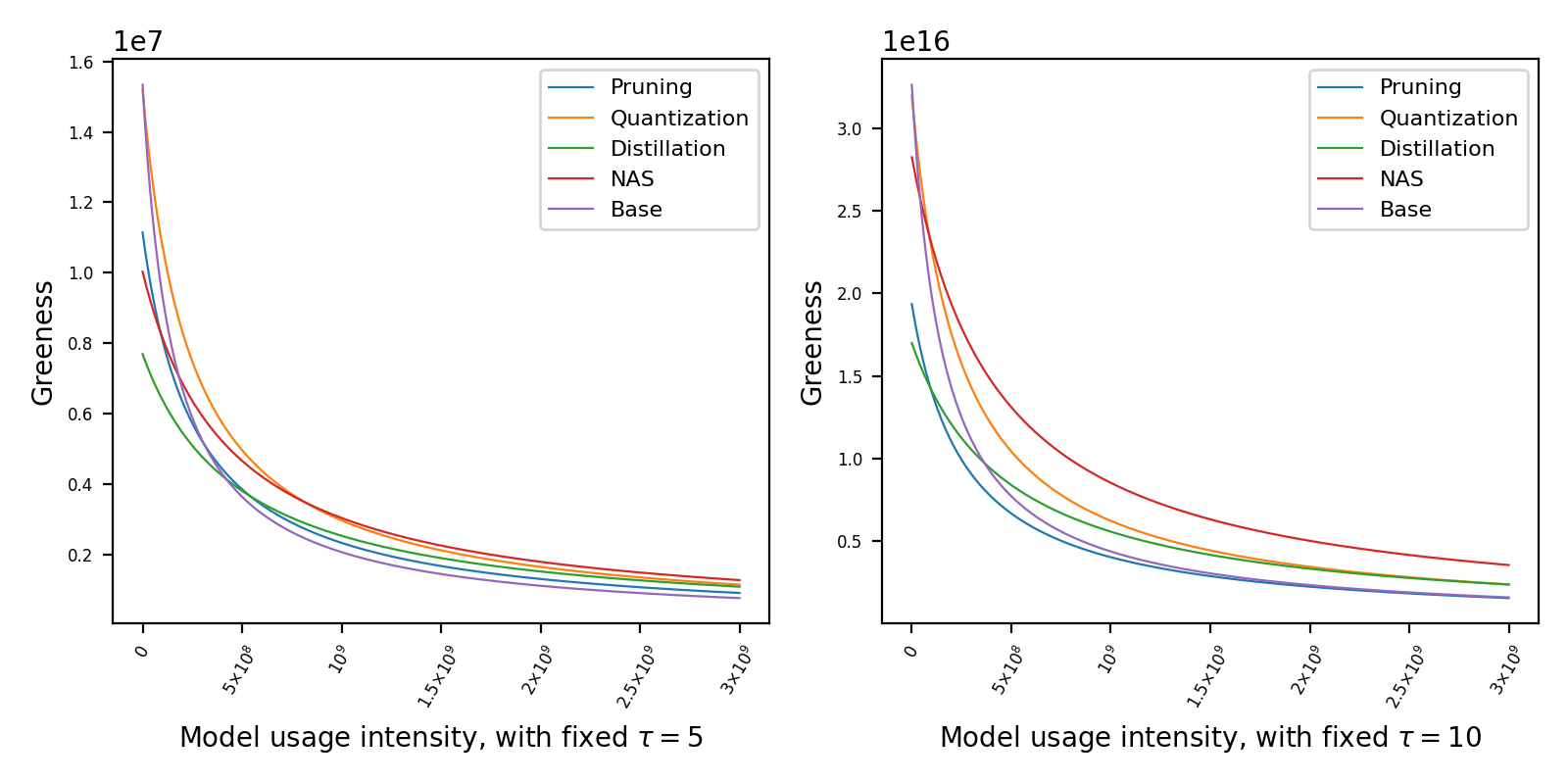}}
    \caption{With several selected algorithms of similar base models, we evaluate the curve of \green{Greeness} as $n$ increases in different $\tau$ value. Y-axis indicates greeness value, X-axis in indicates MUI with $\tau$ is set to 5 and 10.}
    \label{fig:greeness_across_algos}
\end{figure}

\section{Implementation of GPU Tracer}
We read the real-time GPU information during the operation of the algorithm through the interface of \texttt{nvidia-smi}. In the following we will provide an implementation of the GPU Tracer to exactly describe the functionality.

\begin{python}
import re
import subprocess
import threading
import time
import xmltodict

class Tracer(threading.Thread):
    def __init__(self, gpu_num=(0,), sampling_rate=0.1):
        ...

    def run(self):
        while self._running:
            time.sleep(self.sampling_rate)
            self.counters += 1
            results = subprocess.check_output(["nvidia-smi", "-q", "-x"]).decode('utf-8')
            dict_results = xmltodict.parse(results)
            if dict_results['nvidia_smi_log']['attached_gpus'] == '1':
                single_gpu_info = dict_results['nvidia_smi_log']['gpu']
                # read information from 'single_gpu_info'
            else:
                # read information from multiple gpus.
                                         
class GPUTracer:
    all_modes = ['distillation', 'pruning', 'quantization', 'nas']
    is_enable = False

    def __init__(self, mode, gpu_num=(0,), sampling_rate=0.1, verbose=False):
        ...
        
    def wraper(self, *args, **kwargs):
        tracer = Tracer(gpu_num=self.gpu_num, sampling_rate=self.sampling_rate)
        start = torch.cuda.Event(enable_timing=True)
        end = torch.cuda.Event(enable_timing=True)
        start.record()
        tracer.start()
        results = self.func(*args, **kwargs)
        tracer.terminate()
        end.record()
        torch.cuda.synchronize()

        tracer.join()
        ###collect information###
        tracer.communicate()
        if self.verbose:
            print(.....)
\end{python}

Since we implemented this tracer using the decorator feature of Python, it is only necessary to add the corresponding code to the function or code snippet that needs to record, and then the GPU information is recorded at runtime. And in the following we provide the use case of GPU Tracer. We will open-source the code to public at \href{https://dev.azure.com/xinyangjiang/_git/CarbonBenchmarking}{Carbon-Benchmark} (currently only avaiable internally).

\begin{minipage}{\linewidth}
\begin{python}
@GPUTracer(mode='distillation', verbose=True, sampling_rate=0.1)
def train_distill_epoch(...):
    """One epoch distillation"""
    # set modules as train()
    for module in module_list:
        module.train()
    # set teacher as eval()
    module_list[-1].eval()
    ...
    
actual_results, GPU_info = train_distill(...)
# The actual-results is the original returned value of train_distill_epoch
# The GPU_info is the recorded GPU information at runtime. 
\end{python}
\end{minipage}

\section{Full Results of \green{Greeness} Benchmark}
Our benchmark evaluated four types of efficient deep learning algorithms, including (1) distillation. (2) pruning. (3) quantization. (4) neural architecture search. Moreover, the benchmark also includes metrics of basic model training. In total, the benchmark contains 216 sets of experiments across the five types. Due to page limitations, we have temporarily placed the table in \href{https://docs.google.com/spreadsheets/d/1T9-XIrUMAFG7JlbtwNhUjHg4YmXrUDIlUNHf2tEx4Ak/edit?usp=sharing}{full table}, which will be updated later on a website with a more user-friendly way.